\documentclass[preprint,12pt]{elsarticle}

\usepackage{amssymb}
\usepackage{times}
\usepackage{latexsym}
\usepackage{graphicx} 
\usepackage{tabularx} 
\usepackage{booktabs} 
\usepackage{amsmath} 
\usepackage{flushend}  
\usepackage{amssymb} 
\usepackage{enumitem} 
\usepackage{multirow} 
\usepackage[numbers]{natbib}
\usepackage{url}
\usepackage{color, colortbl}
\definecolor{Gray}{gray}{0.9}
\usepackage{microtype}

\newcommand\minilm{\textsc{MiniLM}}
\newcommand\bertbase{BERT$_\text{BASE}$}
\newcommand\meqsum{\textsc{MeQSum}}
\newcommand\bertlarge{BERT$_\text{LARGE}$}

\journal{Journal of Biomedical Informatics}

\begin{document}

\begin{frontmatter}



\title{Question-aware Transformer Models for Consumer Health Question Summarization}
 \author{
Shweta Yadav\corref{cor1}\fnref{fn2}}
    \ead{shweta.shweta@nih.gov}
  \author{Deepak Gupta\fnref{fn2}}
  \ead{deepak.gupta@nih.gov}
     \cortext[cor1]{Corresponding author}
     \fntext[fn2]{Both authors contributed equally to this work.}
    \author{Asma Ben Abacha\corref{cor2}}
 \ead{asma.benabacha@nih.gov}
  \author{Dina Demner-Fushman\corref{cor2}}
 \ead{ddemner@mail.nih.gov}
    \address{LHNCBC, U.S. National Library of Medicine, MD, USA }
   






\begin{abstract}
Searching for health information online is becoming customary for more and more consumers every day, which makes the need for efficient and reliable question answering systems more pressing. An important contributor to the success rates of these systems is their ability to fully understand the consumers’ questions. However, these questions are frequently longer than needed and mention peripheral information that is not useful in finding relevant answers. Question summarization is one of the potential solutions to simplifying long and complex consumer questions before attempting to find an answer. In this paper, we study the task of abstractive summarization for real-world consumer health questions. We develop an abstractive question summarization model that leverages the semantic interpretation of a question via recognition of medical entities, which enables generation of informative summaries. Towards this, we propose multiple Cloze tasks (i.e. the task of filing missing words in a given context) to identify the key medical entities that enforce the model to have better coverage in question-focus recognition. Additionally, we infuse the decoder inputs with question-type information to generate \textit{question-type driven} summaries. When evaluated on the \textsc{MeQSum} benchmark corpus, our framework outperformed the state-of-the-art method by $10.2$ ROUGE-L points. We also conducted a manual evaluation to assess the correctness of the generated summaries. 
\end{abstract}
 


\begin{keyword}
Consumer Health Question Summarization \sep Transformers \sep Abstractive Summarization
\end{keyword}

\end{frontmatter}


\section{Introduction}

Consumers frequently query the web in search of quick and reliable answers to their healthcare information needs. A recent survey showed that on average eight million people in the United States seek health-related information on the Internet every day\footnote{https://pewrsr.ch/3l6m3mv}. A potential solution to support consumers in their health-related information search is to build a natural language question answering (QA) system that can extract both reliable and precise answers from a variety of health-related information sources.
Consumers' questions are often overly descriptive and contain peripheral information such as the patient's medical history, which makes their automatic analysis and understanding more challenging for QA systems. These peripheral details are often not required to retrieve relevant answers, and their removal can lead to substantial improvements in QA performance \cite{abacha2019role}. Hence, novel strategies should be devised for automatic question simplification prior to answer retrieval.

Recent work in abstractive summarization has made significant advancement on large corpora such as news articles \cite{nallapati-etal-2016-abstractive,narayan2018don}, the biomedical literature \cite{zhang2019pegasus}, and clinical records \cite{zhang2019optimizing,macavaney2019ontology}. Two major shortcomings, however, impede higher success in obtaining coherent and fluent summaries for consumer health questions (CHQ): \textbf{(1)} understanding the intent of the medical question and \textbf{(2)} capturing the salient medical entities to ensure a more relevant coverage. In the case of consumer health question answering, two key elements have to be identified correctly: the \textit{question focus} and the \textit{question type} that captures the intent. The question focus is the main entity of the question and the question type is the aspect of interest about the focus. 
 
We can observe from Table~\ref{tab:c2-simple-qg} that many  existing models fail to identify the question focus \textit{``Friedreich's ataxia''} and the original question types: `\textit{Treatment}' and `\textit{Test}'. In these examples, the summarization models generate either a partial question summary (capturing only the `\textit{Treatment}' question type) or an out-of-context summary. These observations indicate that the current models may lack semantic understanding of the input, which is crucial for generating relevant question summaries for question answering. The importance of the question focus and type was specifically highlighted in \citet{Ariel-2017}, where the authors showed that the question focus and the question type are potent factors in retrieving relevant answers. 
\begin{table}[]
\centering
\begin{tabular}{@{}p{\columnwidth}@{}}
\specialrule{.1em}{.1em}{.1em}
\small
\textbf{Original Question} \\
\midrule
\small
I have been told I have Friedreich's ataxia.  I am looking for a treatment to reverse it. I has to do with the chromosome number 9 defective in both of my parents and possible damage to chromosome number 10.  Would you be able to tell me if number 9 is defective in me.
\\
\specialrule{.1em}{.1em}{.1em}
\small
\textbf{Reference Summary} \\
\midrule
\small
where can i get genetic testing for friedreich's, and what are the treatments for it?
\\
\specialrule{.1em}{.1em}{.1em}
\small
\textbf{Generated Summary} \\
\midrule
\small
\textbf{Sequence to Sequence:} what are the treatments for \textcolor{red}{disease}
\\
\small
\textbf{Pointer Generator:} what are the treatments for \textcolor{red}{chromosome chromosome chromosome chromosome [UNK]}
\\
\small
\textbf{\minilm{}}: what are the treatments for \textcolor{red}{chickenrenon's} ataxia ?\\
\specialrule{.2em}{.1em}{.1em} 
\end{tabular}
\caption{Example of generated summaries by existing neural summarization models.}
\vspace{-0.5em}
\label{tab:c2-simple-qg}
\end{table}
 
To this end, we present our framework for abstractive summarization with latent knowledge-induced pretrained language model (LM) in the under-explored area of consumer health question summarization. 
Under the proposed framework, 
we aim to enhance the traditional transformer-based pretrained LM with the knowledge of `question-focus' and `question-type'.\\
\indent We explore two variants of knowledge-induced LM for the task of question summarization. In the first variant (`\textit{question-focus aware}' model), we adapt LM by introducing various Cloze tasks for question summarization. More specifically, we define three Cloze tasks: `\textit{masked seq2seq prediction}', `\textit{masked n-gram prediction}', and `\textit{masked medical entities prediction}' to induce good coverage of medical entities in the summarized question, which might help the model to generate more informative summaries. In the second variant, we infuse the \textit{latent question-type knowledge} into the LM that guides the LM to capture the relevant and appropriate intent of the questions while generating the summary. 





Our proposed models outperformed transformer based pretrained models (e.g. MiniLM) and the state-of-the-art pointer-generator model on the reference \textsc{MeQSum} benchmark for consumer health question summarization \cite{BenAbacha-ACL-2019-Sum}. More importantly, given the small size of the training corpus ($5,155$ samples) when compared to the volume of training corpora in the news articles and biomedical domains, our solution proved effective in harnessing the existing pretrained networks to benefit the task of question summarization. A detailed experimental analysis revealed that the proposed models are capable of generating  questions with significant coverage of the question focus and the question type, leading to relevant and informative question summaries. 

\indent We summarize our contributions as follows:
\begin{itemize}
    \item We propose a simple yet effective method for consumer health question summarization, by introducing various Cloze tasks to pretrained transformer models for better coverage of the \textit{question focus} in the summarized questions.
\item We also introduce an explicit and an implicit ways of infusing the knowledge of the \textit{question type} in pretrained transformer models for the generation of informative and \textit{question-type driven} summaries.
\item Our proposed model achieves the state-of-the-art performance in the consumer health question summarization task and outperforms the recent state-of-the-art pretrained LM models (e.g. T5, PEGASUS, and BART). 
\item We conduct an automatic evaluation on the \meqsum{} benchmark and a manual evaluation to assess the correctness of the generated question summaries.
\end{itemize}

\vspace{-0.4em}

\section{Related Work} 

Abstractive summarization of consumer health questions is at the intersection of three research areas that we discuss in this section. \begin{itemize}
    \item  \textbf{Open-Domain Abstractive Summarization:} Abstractive text summarization approaches made a leap in performance with the development of neural machine translation and text generation models \cite{bahadanau_attention} and the creation of large corpora such as the CNN/Daily Mail dataset \cite{Nallapati-Abstractive-2017}. The most common approaches to  the task of abstractive summarization are sequence-to-sequence (seq2seq) models \cite{Nallapati-Seq2Seq-2016}, and their pointer generator extension \cite{go-to-the-point-see-2017} that addresses out-of-vocabulary words using a copy mechanism. An additional coverage mechanism was also proposed to reduce word repetition.\\
\indent Recently, \citet{lebanoff2020understanding} introduced a `points of correspondence' method based on text cohesion theory to fuse sentences together into a coherent and more faithful summarized text. Few other recent works have exploited question answering and natural language inference (NLI) models to identify factual coherence in the generated summaries \cite{falke2019ranking,kryscinski2019evaluating}. \citet{goodrich2019assessing} compared various models that can extract factual tuples from text utilizing their proposed model-based metrics to analyze the factual accuracy of the generated text. \citet{falke2019ranking} evaluated the factual-correctness of the generated summaries using NLI systems by re-ranking the generated summaries based on the entailment predictions of the NLI models. 
\citet{huang2020knowledge} also exploited open information extraction to extract the salient entities from the document to generate faithful summaries. In another research track, some works explored reinforcement learning (RL) for abstractive summarization by relying on ROUGE scores \cite{paulus2017deep,chen2018fast,dong2018banditsum,pasunuru2018multi} or on multiple RL rewards in an unsupervised setting \cite{laban2020summary}.
\item \textbf{Biomedical and Clinical Domains:} The biomedical domain has also observed a recent surge in the development of neural summarization models. The majority of these models are focused on  summarization of the biomedical literature \cite{cohan2018discourse,zhang2019pegasus} and radiology reports \cite{li2018hybrid,zhang2018learning,zhang2019optimizing}.  \\
\indent For instance, \citet{macavaney2019ontology} improved the  performance of a pointer generator model by augmenting the data with medical ontologies that enable the model to learn and decode the key entities in the generation process. In a similar approach, \citet{sotudeh2020attend} infused a neural summarization model with medical ontology-based terms to solve the content selection problem for clinical abstractive summarization. In a different approach, \citet{cohan2018discourse} developed a discourse-aware neural summarization model that captures the discourse structure of the document via a hierarchical encoder.
\citet{zhang2019pegasus} proposed an encoder-decoder method based on the transformer architecture and on huge text corpora with a novel self-supervised objective function. They evaluated their method on PubMed articles and obtained encouraging ROUGE scores. 

\item \textbf{Consumer Health Questions:} Most of the research work discussed above was conducted on datasets that are often well formed, well structured, and grammatically correct. In contrast, consumer health questions are profuse with misspellings, ungrammatical structures, and non-canonical forms of medical terms that resemble open-domain language more than the medical language. Moreover, the consumer questions submitted to online platforms, such as forums and institutional contact pages, are very often long questions with a substantial amount of peripheral information and multiple sub-questions. This adds to the challenge of developing a summarization model that can account for concise, faithful, and succinct summaries.  \citet{BenAbacha-ACL-2019-Sum} have introduced the consumer-health question summarization task and created a corpus of $5,155$ question summaries. They experimented with seq2seq and pointer-generator models and achieved $44.16$\% ROUGE-1 score in consumer-health question summarization.   

Our work advances the state-of-the-art for the summarization of consumers' questions and introduces new approaches that were specifically designed to preserve the intent of the original questions.  
\end{itemize}
\section{Question Summarization Approach}
 
We tackle the task of consumer health question (CHQ) summarization with the goal of generating summarized questions expressing the minimum information required to find correct answers.
Our proposed summarization models are tailored to generate CHQ summaries that are aware of the question focus and the question type(s). Towards this, we devise `\textit{question-focus aware}' and `\textit{question-type aware}' summarization models. In the `\textit{question-focus aware}' model, we trained the summarization model over three Cloze tasks to ensure the maximum coverage of question focus in the generated summaries. In the `\textit{question-type aware}' summarization model, we guide the model to generate \textit{question-type driven} summaries. We model this by infusing the question type knowledge both explicitly and implicitly in the summarization model. 
 
We use the pretrained language model \minilm{} \cite{wang2020minilm} as the base model architecture to summarize the consumer health questions. We chose \minilm{} for two particular reasons. First, \minilm{} has shown near state-of-the-art results on both natural language generation and understanding. Second, it has fewer parameters ($33$M) compared to other large  transformer-based language models, which mitigates the latency and resource challenges for fine-tuning and for use in real-time applications.

\textsc{MiniLM} is built on deep self-attention distillation framework to compress large Transformer-based pretrained models. In this framework, a small model (student) with fewer parameters is trained by deeply mimicking the self-attention module of the large pretrained model (teacher). The self-attention module of the last Transformer layer of the teacher model is distilled to gather the knowledge to build the student model. Furthermore, \textsc{MiniLM} utilizes the scaled dot-product between values in the self-attention module as the new deep self-attention knowledge, in addition to the attention distributions (i.e., the scaled dot-product of queries and keys) that has been used in the Transformer-based pretrained model. The model is trained by  minimizing the sum of \textbf{(1)} KL-divergence between the self-attention distributions of the teacher and student models, and \textbf{(2)} KL-divergence between the value relation of the teacher and student models.
In the distillation setup, the uncased version of \bertbase{} is used as teacher model. BERT$_\text{BASE}$~\cite{devlin2018bert} has about $109$M parameters with $12$-layer Transformer ($768$ hidden size), and $12$ attention heads. The number of heads of attention distributions and value relation are set to $12$ for \minilm{} student models.
Below we begin by providing details on the input representation followed by our proposed `\textit{question-focus aware}' and `\textit{question-type aware}' summarization models.

\subsection{\textbf{Input Representation}}
\label{sec:input}
 
Given an original question $Q =\{q_1, q_2, \ldots, q_m\}$ and the reference summary ${S}= \{s_1, s_2, \ldots, s_n \}$, we follow the input representation of BERT~\cite{devlin2018bert} and tokenize $Q$ and $S$ using the WordPiece~\cite{wu2016google} tokenizer. Before processing, we also add the \texttt{[CLS]} token to the beginning of the token sequence of the question.

We form a single input token sequence by concatenating the token sequence of the question and their summarized counterpart. We add the conventional separator token (\texttt{[SEP]}) between and at the end of input token sequence. We denote the final input tokens as $\{t_1, t_2, \ldots, t_N\}$, where $N=|Q|+|S|+3$, and $3$ is the sum of one \texttt{[CLS]} and two (\texttt{[SEP]}) tokens.
For each input token, we compute the final token embedding ($\mathbf{v}$) by summing the corresponding token, position, and segment embeddings. The input embeddings can be denoted by the $\mathbf{H}^0 = [\mathbf{e}_1, \mathbf{e}_2,  \ldots, \mathbf{e}_{N}]$.

\subsection{\textbf{Question-Focus Aware Model (QFA)}}
We define multiple Cloze tasks to effectively guide the \minilm{} model towards improved awareness of question focus words for the consumer health question summarization task. Generally, in a Cloze task, some tokens are chosen randomly and replaced with special token (\texttt{[MASK]}), which is predicted during training.
We consider each medical entity as a candidate for the question focus. 
\paragraph{\textbf{Cloze Task 1 - Sequence-to-Sequence Masking}}
\label{sec:s2s:lm}
We adapt the seq2seq LM objective function as discussed in \cite{unilm}. We aim to predict the masked token from the reference summary using the original question token sequence as input and the reference summary tokens which are to the left\footnote{While generating at a given time step $t$, the model only has the access to the $<t$ tokens.} of the masked token and current token. For example, given question tokens $t_1^{q}, t_2^{q}, t_3^{q}$ and its reference summary tokens $t_4^{s}, t_5^{s}, t_6^{s}$, we feed the input token sequence as ``\texttt{[CLS]}, $t_1^{q}, t_2^{q}, t_3^{q}$, \texttt{[SEP]}, $t_4^{s}, t_5^{s}, t_6^{s}$, \texttt{[SEP]}'' to the model. The summary token $t_5^{s}$ has access to the first seven tokens i.e. all question tokens (including \texttt{[CLS]} and \texttt{[SEP]}), and the previous tokens in the reference summary, and current token.
This Cloze task objective mimics the seq2seq generation paradigm, where the encoder encodes the entire source sentence and makes it available to the  decoder that generates one word at a time, given the encoder states and all the previously generated words.

The input vector $\mathbf{H}^0$ is passed to $L$-size stack of Transformer block \cite{vaswani2017attention} to obtain the contextual representations of each token. Formally, at a given level $l$, we compute the contextual representation as follows:
\begin{equation}
    \label{transformer}
    \small
    \mathbf{H}^l = \texttt{Transformer}_{l}(\mathbf{H}^{l-1}), l \in [1,L].
\end{equation}

For the $l^{th}$ Transformer layer, we compute the query ($\mathbf{\hat{Q}}$), key ($\mathbf{\hat{K}}$), and value ($\mathbf{\hat{V}}$) vectors by projecting the output of the $(l-1)^{th}$ Transformer layer as follows:
\begin{equation}
\small
    \label{query-key-value}
    \mathbf{\hat{Q}} = \mathbf{H}^{l-1} \mathbf{W}_l^{\hat{Q}} ,\quad \mathbf{\hat{K}} = \mathbf{H}^{l-1} \mathbf{W}_l^{\hat{K}} ,\quad \mathbf{\hat{V}} = \mathbf{H}^{l-1} \mathbf{W}_l^{\hat{V}}
\end{equation}

where the previous layer's output $\mathbf{H}^{l-1} \in \mathbb{R}^{N \times d_h}$, parameter matrices $\mathbf{W}_l^{\hat{Q}} , \mathbf{W}_l^{\hat{K}} , \mathbf{W}_l^{\hat{V}} \in \mathbb{R}^{d_h \times d_k}$, and $d_h$ and $d_k$ are the dimensions of the hidden state and scaling factor respectively. 

In order to facilitate the token access behaviour, we compute the output of a self-attention head  $\mathbf{A}_l$ as follows: 
\begin{equation}
\label{attention-head}
\small
\begin{split}
\mathbf{A}_l &= softmax(\frac{\mathbf{{\hat{Q}}} \mathbf{{\hat{K}}}^{\intercal}}{ \sqrt{d_k}} + \mathbf{M}) \mathbf{{\hat{V}}}_l\\
\small
  \mathbf{M}_{ij} &= \begin{cases} 0, &\text{allow the $j^{th}$ token to attend $i^{th}$} \\ -\infty, &\text{otherwise} \end{cases} 
  \end{split}
\end{equation}

where the mask matrix $\mathbf{M} \in \mathbb{R}^{N \times N}$  determines whether a pair of tokens can be attended to each other. 
We create the mask matrix $\mathbf{M}$ for each pair of $Q$ and $S$. 
For each token $t_j$ of $S$, we set the appropriate value (based on Eq. \ref{attention-head}) in mask matrix $\mathbf{M}$ such that, it only attends to all the tokens which are to the left of $t_j$ including self. 
The final hidden state for the $i^{th}$ input token is computed from the output of the last Transformer layer ($H^{L} \in \mathbb{R}^{N \times d_h}$) as $h_i \in \mathbb{R}^{d_h}$. We transform the hidden state to the vector of the vocabulary dimension as follows:
\begin{equation}
\label{prediction}
\small
\begin{split}
    T &= gelu(H^{L} \mathbf{W} + \mathbf{b}_{T})\\
    P & = softmax(T \mathbf{U} + \mathbf{b}_{P}) 
\end{split}
\end{equation}
where, $\mathbf{W} \in \mathbb{R}^{d_h \times d_h}$ and $\mathbf{U} \in \mathbb{R}^{d_h \times |V|}$ are the weight matrices and $\mathbf{b}_T \in \mathbb{R}^{d_h}$ and $\mathbf{b}_P \in \mathbb{R}^{|V|}$ are the bias terms associated with the transformed hidden state $T \in \mathbb{R}^{N \times d_h}$ and prediction $P \in \mathbb{R}^{N \times |V|}$ respectively. $|V|$ denotes the vocabulary size.
We minimize the cross-entropy loss to train the network. Formally,
\begin{equation}
    \label{cross-entropy-loss-s2s}
    \small
    \mathcal{L}_{s2s} = -\sum_{i=1}^{|V|} y_i log p_i
\end{equation}
where, $y_i$ is the actual token probability of the $i^{th}$ token and $p_i$ is the predicted token probability.
We followed the masked word prediction similar to \cite{devlin2018bert} and  choose the $i^{th}$ token positions from the summary sentence at random for prediction. We replaced the $i^{th}$ token \textbf{(a)} with the \texttt{[MASK]} token $80$\% of the time \textbf{(b)} a random token from vocabulary $10$\% of the time \textbf{(c)} the unchanged $i^{th}$ token $10$\% of the time. 
\paragraph{\textbf{Cloze Task 2 - N-gram Masking}} As many medical/clinical entities  are multi-word expressions, the traditional \textsc{MiniLM} masking strategy may not be useful to correctly predict the medical entities. To address this, we introduce n-gram masking, which randomly selects an n-gram (from $1$ to $3$) from the summary sentence to mask and predict them using the given consumer health question similar to the Cloze Task 1 objective. We compute the cross-entropy loss ($\mathcal{L}_{ngm}$) similar to the Eq. \ref{cross-entropy-loss-s2s}.
\paragraph{\textbf{Cloze Task 3 - Medical Entity Masking}} 
We design another Cloze task objective to train the summarization network where we first identify the medical entities\footnote{We utilize the Scispacy \texttt{en\_ner\_bionlp13cg\_md} model trained on the  BioNLP13CG dataset \cite{pyysalo2015overview} to identify medical entities belonging to $16$ distinct UMLS semantic types.}.
This important LM objective enhances the model's ability to generate medical terms correctly based on the given consumer health question. Predicting the correct medical entities leads to more relevant question summaries. We compute the cross-entropy loss ($\mathcal{L}_{fwm}$) similar to the Eq. \ref{cross-entropy-loss-s2s}.
\paragraph{\textbf{Multi-Cloze Fusion}} We utilize the benefits of all the aforementioned Cloze tasks and train the \textit{question-focus aware} summarization model by minimizing the aggregated (sum) loss computed from each Cloze task. Formally, 
\begin{equation}
\small
    \label{qfa_loss}
    \mathcal{L}_{qfa} =   \mathcal{L}_{s2s} +   \mathcal{L}_{ngm} + \mathcal{L}_{fwm}
\end{equation}
where, $\mathcal{L}_{qfa}$ is the loss of the \textit{question-focus aware} summarization model. 
\paragraph{Testing Phase.} Once the model is trained on the Cloze tasks, it can be used for testing/generating summaries. The model uses all the tokens from the original question $Q$ and previously generated sequence until the $t-1$ time step to generate at a given time step $t$. This process is continued until the generated text has the length of the target summary. We utilize Beam Search to generate optimal summary with a beam size of five. 
\subsection{\textbf{Question-Type Aware Model (QTA)}}
We incorporated the question-type knowledge in two ways: \textit{explicit} and \textit{implicit} knowledge infusion. We began by first identifying the question types followed by their infusion in the summarization model. 
\paragraph{\textbf{Question Type Identification}} 
Following a previous study of consumer health questions \cite{kilicoglu2018semantic}, we use a heuristic to categorize the original question $Q$ to one of the question-types (\textit{Information}, \textit{Treatment}, \textit{Testing}, \textit{Cause}, \textit{Physician}, \textit{Ingredients}, and \textit{Other}). 
We searched for the specific question type with its lemma word in the lemmatized reference summary and labeled each consumer question.  Since, the reference summary will not be available at the inference time, we aim to map the question type with the original question $Q$ and infuse the question-type knowledge into the question summarization model. Specifically, in the explicit knowledge infusion, we utilize the \bertlarge{} model with two layers of feed-forward networks to predict the question type given $Q$ at test stage. In the implicit question-type knowledge infusion, we used a multi-task learning (MTL) approach to simultaneously predict the labels and generate the question summary.\\
\paragraph{\textbf{1. Explicit QTA Knowledge Infusion}}

In the explicit knowledge infusion method, we aim to integrate the question-types information directly into the question summarization network. 
We learn a weight matrix $Q_E \in \mathbb{R}^{|C| \times d_h}$ to encode the question-type information associated with a particular question-type in the summarization model. We augment the predicted $j^{th}$ question-type vector $Q_E[j]$ with the final hidden state $h_i \in \mathbb{R}^{d_h}$ for the $i^{th}$ input token. Formally,
\begin{equation}
\label{explicit-augumentation}
\small
\begin{split}
h_i^{E} = tanh ( h_i + Q_E[j])
\end{split}
\end{equation}
where, $h_i^{e} \in \mathbb{R}^{2d_h}$ is question-type augmented hidden state for the $i^{th}$ token. Thereafter, we follow Eq. \ref{prediction} to predict the masked token and use Eq. \ref{cross-entropy-loss-s2s} to compute the training loss for the question summarization model.
\paragraph{\textbf{2. Implicit QTA Knowledge Infusion}} In the first explicit knowledge infusion method, we utilized the pre-trained network to capture question-type semantics. However, two main challenges need to be addressed in the explicit question-type knowledge infusion: \textbf{(1)} its high dependency on the performance of the underlying classifier, and  \textbf{(2)} the difficulty to learn the mapping function $f:Q \rightarrow C$ that maps a lengthy question ($Q$) having multiple sub-questions to a question type ($C$). To tackle these challenges, we attempt to model question-type knowledge infusion implicitly in the summarization model. 

We propose a multi-task learning approach, where the model learns to recognize the question-type semantics via the auxiliary task of `\textit{Question-Type Prediction}' along with the primary task of `\textit{Question Summarization}'. We use the shared encoder from the \minilm{} to encode the consumer health questions.  
Formally, we classify each question into one of the seven aforementioned categories as:
\begin{equation}
\label{implicit-q-type}
\small
\begin{split} 
    h_I &= tanh(h_0 \mathbf{W_I} + \mathbf{b}_{h})\\
    C_I & = softmax(h_I \mathbf{U_I} + \mathbf{b}_{c})
\end{split}
\end{equation}

 The training on the  \textit{`Question-Type Prediction'} is performed by minimizing the cross entropy loss ($\mathcal{L}_{qtype}$). Similar to the explicit question-type knowledge infusion (Eq. \ref{explicit-augumentation}), we augment the learned question-type vector $h_I$ with the hidden state $h_i \in \mathbb{R}^{d_h}$ for the $i^{th}$ input token and computed the final hidden state as $h_i^{I}$. We train the network by minimizing the following loss function:
\begin{equation}
    \label{mtl_loss}
    \mathcal{L}_{model} =   \mathcal{L}_{s2s} +   \mathcal{L}_{qtype}
\end{equation}

\section{Experimental Results \& Analysis}
 
\subsection{\textbf{Datasets}}   \label{sec:dataset}
We used the summarization dataset of consumer health questions \meqsum{}, consisting of $1,000$ consumer health questions and their corresponding summaries. We followed the same data augmentation method as \citet{BenAbacha-ACL-2019-Sum}, and added a set of $4,655$ pairs of clinical questions asked by family doctors and their short versions collected from \citet{ely2000taxonomy}. We performed all the experiments\footnote{Wherever we refer to the \meqsum{} dataset, we refer to the augmented dataset having $5,155$ training pairs and $500$ test samples.} with $5,155$ training pairs and $500$ test samples pairs.

For question-type identification, we used a dataset of consumer health questions   \cite{kilicoglu2018semantic}  that contains two types of questions: CHQA-web (MedlinePlus\footnote{\url{https://medlineplus.gov/}} queries) and CHQA-email (questions received by the U.S. National Library of Medicine\footnote{\url{https://www.nlm.nih.gov/}}), consisting of $23$ and $31$ question types, respectively.  

\subsection{\textbf{Hyper-parameters}}
We use a beam search algorithm with a beam size of $5$ to decode the summary. We train all summarization models on the training dataset for $20,000$ steps, with a batch size of $16$. We set the maximum question and summary length to $100$ and $20$, respectively. To update the model parameters, we use the Adam \cite{adam} optimization algorithm with a learning rate of $7e-5$. The hidden size of \minilm{} is $384$; we set the hidden vector size to $384$ in all experiments. We use a linear learning rate decay schedule, where the learning rate decreases linearly from the initial learning set in the optimizer to $0$. To avoid the gradient explosion issue, the gradient norm was clipped within $1$.

\subsection{\textbf{Baseline Models}}
We experimented with several competitive baseline models:  Seq2Seq, Pointer Generator, and pretrained transformers models (Transformers, Presumm, T5, PEGASUS, BART, MiniLM) as shown in Table~\ref{tab:result}. Below we explain each of the baseline models in detail.

\begin{table*}[]
\resizebox{\linewidth}{!}{%
\begin{tabular}{c|c|cccccccccc}
 \hline
\multicolumn{2}{c|}{\multirow{2}{*}{\textbf{Models}}} & \multicolumn{1}{c}{\textbf{ROUGE-1}} & \multicolumn{1}{c}{\textbf{ROUGE-2}} &
 \multicolumn{1}{c}{\textbf{ROUGE-L}} & 
 \\ \cline{3-6} 
\multicolumn{2}{c|}{}  &
\textbf{F1}  &
\textbf{F1} 
&\textbf{F1} \\ \hline
\multirow{5}{*}{\textbf{Baselines}} & \begin{tabular}[c]{@{}c@{}} Seq2Seq \cite{seq2seq}\\  \end{tabular}    
&25.28  &14.39  &24.64 \\ 
 & Seq2Seq + Attention \cite{bahadanau_attention}    
 & 28.11    & $17.24$   &$27.82$ \\
  & Pointer Generator (PG) \cite{go-to-the-point-see-2017}   & 32.41  & 19.37  & 36.53 \\ 
  & \begin{tabular}[c]{@{}c@{}}SOTA \cite{BenAbacha-ACL-2019-Sum}  \end{tabular}  & 40.00 & 24.13  & 38.56 \\ \hline
  \multirow{4}{*}{ \begin{tabular}[c]{@{}c@{}} \textbf{Baselines-Transformers} \end{tabular}}
 & Transformer \cite{vaswani2017attention}  & 25.84   & 13.66   & 29.12 \\
 & PreSumm \cite{liu-lapata-2019-text}  & 26.24   & 16.20   & 30.59 \\ 
 & T5 \cite{raffel2019exploring}  & 38.92   & 21.29   & 40.56 \\ 
 & PEGASUS \cite{zhang2019pegasus} & 39.06   & 20.18   & 42.05 \\
 & BART \cite{lewis2019bart}   & 42.30   & 24.83   & 43.74 \\ 
 & \minilm{} \cite{wang2020minilm} (\textbf{M1}) & 43.13   & 26.03   & 46.39 \\ \hline
 \begin{tabular}[c]{@{}c@{}} \textbf{Proposed Method-I (QFA)} \\ \end{tabular}
 & Multi-Cloze Fusion (\textbf{M2})  & 44.58  & 27.02  &47.81 \\ \hline
 \multirow{2}{*}{\begin{tabular}[c]{@{}c@{}} \textbf{Proposed Method-II (QTA)} \\  \end{tabular}}
 & Explicit QTA Knowledge-Infusion (\textbf{M3})  & \textbf{45.20}   & \textbf{28.38}  & \textbf{48.76} \\ 
 & Implicit QTA Knowledge-Infusion (\textbf{M4})  & 44.44   & 26.98  & 47.66 \\ \hline
\end{tabular} 
} 
\caption{Performance comparison of the proposed models and various baseline models on the MeQSum dataset. The highlighted rows represent the best results obtained by the proposed method (M3).}
\label{tab:result}
\end{table*} 

\begin{itemize}
    \item \textbf{Seq2Seq} \citep{seq2seq}: This is the first baseline  on summarizing consumer health questions. We use a one-layer LSTM with a hidden dimension of  $512$ for both the encoder and the decoder. 
   \item  \textbf{Seq2Seq + Attention} \citep{bahadanau_attention}: This baseline is the extension of the  \textit{Seq2Seq} model with the attention mechanism. 
    \item \textbf{Pointer Generator (PG)} \cite{go-to-the-point-see-2017}: This model  extends the \textit{Seq2Seq + Attention} baseline model with the copy mechanism to handle the rare and unknown words.
    \item \textbf{PG + Coverage} \cite{BenAbacha-ACL-2019-Sum}: We also compare our proposed model with the current state-of-the-art on consumer health question summarization. This model is the extension of the PG model with the coverage mechanism \cite{go-to-the-point-see-2017} to avoid the generation of repeated words in the summary. 
    
    \item \textbf{Transformer} \cite{vaswani2017attention}: We use the Transformer encoder-decoder architecture to train the network with the training dataset presented in Section \ref{sec:dataset}. Both of our encoder and decoder consist of six  Transformer blocks.    
    \item \textbf{PreSumm} \cite{liu-lapata-2019-text}: 
    We use the PreSumm architecture proposed in \cite{liu-lapata-2019-text} to train the question summarization network. It is a  standard encoder-decoder framework for abstractive summarization, where the encoder is a pretrained BERTSUM \cite{liu-lapata-2019-text} and the decoder is a 6-layered Transformer. 
\item \textbf{T5} \cite{raffel2019exploring}: This is another pre-trained model developed by exploring the transfer learning techniques for natural language processing (NLP) by introducing a unified framework that converts all text-based
language problems into a text-to-text format. The T5 model is an encoder-decoder Transformer with some architectural changes as presented in detail in \cite{raffel2019exploring}. 
\item \textbf{PEGASUS} \cite{zhang2019pegasus}: It is a large Transformer-based encoder-decoder model pre-trained on massive text corpora with a novel self-supervised objective, called Gap Sentences Generation, specially designed to pre-train the Transformer model for abstractive summarization. The important sentences from the document are masked and are generated together as one output sequence from the remaining sentences of the document. 
\item \textbf{BART} \cite{lewis2019bart}:  It is a denoising autoencoder model
for pre-training sequence-to-sequence models.
BART is trained by (1) corrupting text with an arbitrary noising function, and (2) learning to reconstruct the original text. 
We fine-tuned the BART model on the training dataset presented in Section \ref{sec:dataset}.

    \item \textsc{\textbf{MiniLM}} \cite{wang2020minilm}: We compare our proposed model to the Transformer-based pre-trained model for summarization, which has shown promising results on  abstractive summarization and question generation tasks. For the Transformer based model, we fine-tune \textsc{MiniLM} on the \textsc{MeQSum} dataset. Thereafter, the fine-tuned \textsc{MiniLM} model is used to summarize the consumer health questions.
    \end{itemize}


\subsection{\textbf{Results}} 
We evaluated the performance of the summarization models using the standard ROUGE metric \cite{lin2004rouge} and reported the ROUGE-1, ROUGE-2, and ROUGE-L\footnote{We use the following ROUGE implementation:  \url{https://pypi.org/project/pyrouge/0.1.3/}.}. Table-\ref{tab:result} provides an overview of the results, which demonstrates that both the `\textit{QFA}' and the `\textit{QTA}' models perform better than all the baseline models including the transformer-based pretrained networks. Our proposed `\textit{QFA}' model achieves a ROUGE-L (F1) score of $47.81\%$, which is $1.42\%$ better than the \textsc{MiniLM} model. Similarly, our proposed `\textit{QTA}' model achieves a ROUGE-L (F1) score of $48.76\%$, which is $2.37\%$ better than the \textsc{MiniLM} model. In comparison to the model developed by \cite{BenAbacha-ACL-2019-Sum} on the \textsc{MeQSum} dataset, we obtained a substantial improvement of $9.25\%$ and $10.2\%$ with the `\textit{QFA}' and `\textit{QTA}' models, respectively. We further provide an ablation analysis (in Table \ref{tab:result-CLOZE-task}) of the question-focus aware summarization model by training the network only with a particular Cloze task.

\subsection{\textbf{Discussion}} 
The results support two important claims: \textbf{(1)} Infusing the knowledge of \textit{question types} leads to more informative question summaries, and \textbf{(2)} Awareness of salient medical focus words in the learning process enables the generation of relevant summaries.
The ablation analysis reveals that the system trained with the Cloze Task 1 objective performs comparatively better than the rest of the objectives. It can be explained by the fact that the Cloze Task 1 objective is to mask a word  from the summary sentence, whereas in the second and third Cloze Tasks the phrases are masked. It is more challenging to predict the entire masked phrase compared to a word. However, the fusion of all the Cloze task objectives complements each other and improves the performance of the summarization system in terms of ROUGE-L (F1) score ($47.81$). 

 \begin{table}[ht]
\centering
\resizebox{0.7\linewidth}{!}{%
\begin{tabular}{c|ccc}
\hline
\multirow{2}{*}{\textbf{Models}} & \textbf{ROUGE-1} & \textbf{ROUGE-2} &
 \textbf{ROUGE-L} \\ \cline{2-4} 
&\textbf{F1} & 
\textbf{F1} & 
\textbf{F1} \\ \hline
 \begin{tabular}[c]{@{}c@{}} Cloze Task 1 \\  \end{tabular} & $43.81$   & $26.6$   & 46.97    \\ 
 Cloze Task 2 & 43.13 &26.7 &46.14 \\ 
  Cloze Task 3 & 44.58  & 27.02  & 47.81 \\  \hline
 \hline
\end{tabular}
}
\caption{Ablation study of various Cloze tasks on the MeQSum dataset.}
\label{tab:result-CLOZE-task} 
\end{table}

We also extend our studies to analyze the effect of various question-type knowledge sources. Towards this, we utilized the Consumer Health Question Answering (CHQA) datasets created by \citet{kilicoglu2018semantic}. The CHQA dataset provides the detailed annotations for the question focus and question types. It consists of two sub-datasets, CHQA-web (MedlinePlus\footnote{\url{https://medlineplus.gov/}} queries) and CHQA-email (questions received by the U.S. National Library of Medicine\footnote{\url{https://www.nlm.nih.gov/}}).  The CHQA-web has $23$ different question types, while CHQA-email has $31$ distinct question types.  \\
\indent We performed an experiment to analyze the effect of different question types obtained from training the Question Type Identification model (\bertlarge{}) on CHQA-web and CHQA-email datasets. The results shows that fine-tuning the (\bertlarge{}) model with heuristically predicted question-types performs better than CHQA-web and CHQA-email dataset with implicit QTA summarization model. This may be because, both CHQA-web and CHQA-email contain very few samples ($874$ samples for CHQA-web and $554$ samples for CHQA-email) for fine-tuning.

We also perform an upper-bound analysis by replacing the predicted question type with the gold-standard question type for the question set of the \meqsum{} dataset. We reported the  performance of the summarization model  with the gold-standard question type in Table \ref{tab:result-question-type}. The system achieves a ROUGE-L score of $54.03$, which strengthens our hypothesis that knowing the question type benefits the summarization model. 
 

\begin{table}[ht]
\resizebox{\linewidth}{!}{%
\begin{tabular}{c|ccc}
\hline
\multirow{2}{*}{\textbf{Question Types Source}} & \textbf{ROUGE-1} & \textbf{ROUGE-2} &
 \textbf{ROUGE-L} \\ \cline{2-4} 
&\textbf{F1} & 
\textbf{F1} & 
\textbf{F1} \\ \hline
 \begin{tabular}[c]{@{}c@{}} CHQA-web \\  \end{tabular} & $44.06$   & $27.29$   & 47.39    \\ 
CHQA-email & 45.02 &27.27 &48.01 \\ 
Heuristically-predicted Question Type & 45.20  & 28.38  & 48.76 \\  
 \hline
Gold-standard question type (upper bound) & 50.45 &32.82 &54.03 \\ 
 \hline

 \hline 
\end{tabular}
}
\caption{Performance comparison of the  \textit{`Question-type aware summarization model'} on different question types from the CHQA-web and CHQA-email datasets.}
\label{tab:result-question-type}
\end{table}

\subsection{\textbf{Evaluation of the Proposed Summarization Model for Question Answering} }
In addition to our experiments on question summarization, we also performed experiments on the QA task to assess whether the summarized questions generated from the proposed summarization model can help improving the QA performance. Towards this, we utilize the LiveQA 2017 test dataset \cite{Abacha2017OverviewOT}, that consists of $104$ medical questions from the National Library of Medicine (NLM). The QA task consists of retrieving a correct answer for each medical question. 
  
We generated the summaries of the test questions using our proposed summarization approach (Model M3) and developed a QA method based on an  Information Retrieval (IR) model to retrieve relevant/similar questions and their associated answers from a collection of question-answer pairs. We utilized the search engine Terrier \cite{ounis06terrier-osir}  and the MedQuad collection \cite{BenAbacha-BMC-2019} to retrieve relevant answers for the summarized and original questions. The MedQuad collection contains 47k medical questions and their answers. We indexed the MedQuad questions and used the TF-IDF model to retrieve questions that are relevant to the original/summarized test questions.

We evaluated the answers of the retrieved questions for the LiveQA test questions using the publicly available judgments\footnote{\url{github.com/abachaa/MedQuAD (QA-TestSet)}}. For the answers that were not judged earlier ($25$ answers), two annotators (a medical
doctor and a medical informatics expert) evaluated them manually and provided their judgement scores following the LiveQA'17 reference answers \cite{Abacha2017OverviewOT}. 
For a fair comparison, we also used the same judgment scores as established by the LiveQA shared task: ``Correct and Complete Answer"
(4), ``Correct but Incomplete" (3), ``Incorrect but Related" (2) and ``Incorrect" (1).

\paragraph{\textbf{QA Results}} To evaluate the QA performance using the original vs. summarized questions, we utilized the evaluation metrics proposed by the LiveQA shared task (these metrics evaluate the first retrieved answer for each test question): 

\begin{itemize}   
\item avgScore(0-3): the average score over all questions, transferring 1-4 level
grades to 0-3 scores. This is the main score to rank the LiveQA systems.  
\item succ@k: the number of questions with score k or above (k= \{2, 3, 4\}) divided by the total number of questions.  
\item prec@k:  the number of questions with score k or above (k= \{2, 3, 4\}) divided by the number of questions answered by the system.
\end{itemize} 

Table~\ref{table:IR} presents the results obtained by the IR-based QA system using: (i) the original questions, (ii) the automatically summarized questions by our proposed approach, and (iii) manually created reference summaries as reported in \cite{abacha2019role}. 

\begin{table*}[ht]
\resizebox{\linewidth}{!}{%
\begin{tabular}{l|c|c|c}
\hline
\textbf{Measures} &
  \textbf{\begin{tabular}[c]{@{}l@{}}Original Questions\end{tabular}} &
  \textbf{\begin{tabular}[c]{@{}c@{}}Summarized Questions \\ (using our proposed approach)\end{tabular}} &
  \textbf{\begin{tabular}[c]{@{}c@{}}Reference Summaries \\ (Manual)\end{tabular}} \\ \hline
avgScore(0-3) & 0.673 & 0.875 & 1.125 \\ \hline
succ@2+       & 0.403 & 0.567 & 0.663 \\ \hline
succ@3+       & 0.201 & 0.23  & 0.317 \\ \hline
succ@4+       & 0.067 & 0.076 & 0.144 \\ \hline
prec@2+       & 0.424 & 0.567 & 0.663 \\ \hline
prec@3+       & 0.212 & 0.23  & 0.317 \\ \hline
prec@4+       & 0.07  & 0.076 & 0.144 \\ \hline
\end{tabular}
}
\caption{Evaluation of the answers retrieved using the original, automatic, and manual summaries based on the LiveQA metrics. }
\label{table:IR}
\end{table*}

The results show that summarizing the consumer health questions can improve the performance of the IR/QA system in retrieving relevant answers. We also observe that the performance of the IR/QA model using the automatically summarized questions by our proposed approach is close to the performance achieved using the manually created reference summaries. 
\subsection{\textbf{Manual Evaluation}} 

To study the correlation between the automatic (ROUGE-1) and human evaluations in question summarization, we randomly selected $50$ questions from the test set and manually evaluated the summaries generated by the best baseline model (M1) and the three proposed models (M2, M3, and M4).  Three annotators experts in medical informatics and medicine performed the analysis on the randomly selected $50$ questions.

We used three scores: $0$ (incorrect summary), $1$ (acceptable summary), and $2$ (perfect summary) to judge the correctness of the generated summaries. 
For each model, we aggregate the scores assigned to each generated summary. We present the normalized manual evaluation scores in Table~\ref{tab:ManualVal}. We also compute the  inter-annotator agreement (IAA) using the Fleiss' kappa \cite{fleiss1971measuring} and report the agreement of $72.67$, $71.09$ and $67.39$ for model M2, M3 and M4 respectively.

The manual evaluation reveals a significant decline in performance of the model (M1) ($18$ points less, on average) over the other proposed models, while the decline in performance according to ROUGE was only $1.5$ points less. 
Our analysis also shows that the proposed question focus-aware summarization model (M2) can generate more relevant and factually correct summaries due to its better coverage in identifying the question focus.
We also observed that model (M3) with the explicit knowledge of the question types is more capable of generating near perfect summaries than the other models.

\begin{table}[]
\centering
\resizebox{0.9\linewidth}{!}{%
\begin{tabular}{ccccccc}
\hline
\multirow{2}{*}{\textbf{Method}} & \multicolumn{2}{c}{\textbf{Evaluation}} & \multicolumn{4}{c}{\textbf{Error Distribution}} \\ \cline{2-7}
                        & Manual        & Automatic       & E1       & E2       & E3      & E4      \\
 \hline  
 M1 & 24 & 42.02 & 8.19 & 8.19 & 31.14 & 42.62 \\
 M2 & 46 & 53.64 & 11.32 & 9.43 & 28.30 & 26.41\\  
 M3 & 38 & 49.65 & 16.66 & 7.40 & 29.62 & 22.22 \\  
 M4 & 42 & 51.20 & 20.68 & 8.62 & 31.03 & 17.24 \\ \hline  
\end{tabular}    
}
\caption{Comparison between the manual and automatic (ROUGE-1) evaluations and the error distributions of the generated summaries on 10\% of the test set. All numbers are in percentage (\%).}     
\label{tab:ManualVal} 
\end{table}

\subsection{\textbf{Error Analysis}}

 We carried out an in-depth analysis of the generated summaries by the models M1, M2, M3, and M4. Table~\ref{tab:Error} presents examples of generated errors by the summarization models. The error distribution for each model is also reported in Table \ref{tab:ManualVal}. We identified different sources of errors as follows:

 \begin{itemize}
\item \textbf{E1 - Partial Question Types.} This is one of the common errors made by the models. If there are more than one question type in the original question, the models fail to generate the complete summary covering all the questions types. 
\item \textbf{E2 - Semantic Inconsistency.} We observe that the models sometimes fail to understand the semantic meaning of the focus words, and therefore generate wrong question types for a given focus category. For example: ``\textit{What is a treatment for Cetrazine}". ``Cetrazine" is a drug name, the model misunderstood it as a medical problem.
\item \textbf{E3 - Incorrect Question Type.} Our analysis on question-type aware model (M3) reveals that though our model handles the question types correctly to certain extent, there are some instances where due to the lack of model inference capability, the models generate summaries with wrong question types.
\item \textbf{E4 - Partial or Incorrect Focus Words.} We observe that sometimes models fail to identify the exact focus words leading to incorrect summaries. For example, in Table-\ref{tab:Error} Question \#3, the model only identified `\textit{poison}' instead of `\textit{mercury poisoning}'. 
\end{itemize}

 \begin{table*}[h]              
\centering 
\begin{tabular}{p{2.3cm}|p{10cm}} \hline     
 \textbf{Question \#1} & \small{ClinicalTrials.gov - Question - specific study. my son is 17 years old he has omental panniculitis (weber cristian disease) can your institute help me to suggest what is this disear and how it care   if you need i can send the full detail}  \\ \hline 
\textbf{\small{Reference}} & \small{Where can I find information on weber christian disease, including treatment for it?}  \\  
 \textbf{\small{Error-Type1}} & \small{where can i find information on weber christian disease ?} \\ 
 \textbf{\small{Error-Type2}} & \small{what is the latest research on weber christian disease ?} \\ \hline 
  \hline 
\textbf{Question \#2} & \small{SUBJECT: pregnancy MESSAGE: I just want to ask if have some ways or have a operation to be pregnant a woman like me had a tubal ligation before  7yrs. Ago.}  \\ \hline 
\textbf{\small{Reference}} & \small{Where can I find information on tubal ligation reversal?}  \\  
 \textbf{\small{Error-Type3}} & \small{what are the treatments for tubal ligation ?} \\ \hline
 \hline 
\textbf{Question \#3} & \small{MESSAGE: Hello, about 9 years ago I was poisoned and I didn't know what it was that was fed to me until a month ago. It was mercury and I'm losing my vision among other illnesses and don't know where to turn for help.}  \\ \hline 
\textbf{\small{Reference}} & \small{What are the treatments for mercury poisoning?}  \\ 
 \textbf{\small{Error-Type4}} & \small{what are the treatments for poison in a 9 - years - old ?} \\ \hline 
\end{tabular}   
\caption{Examples of various types of errors in the summaries generated by the M1-4 models.}       
\label{tab:Error} 
\end{table*}  

\section{Conclusion} 
In this work, we tackle the summarization of long and complex  consumer health questions. We propose question-focus and question-type aware summarization models that are able to generate relevant and succinct summaries of the original questions. Experiments show that our proposed model achieves state-of-the-art performance on the \meqsum{} dataset outperforming various encoder-decoder and existing pretrained transformer-based summarization models. We also explore the use of the automatically generated summaries in an IR-based QA system and find that the automatic summaries improve the QA performance. 
The two proposed question summarization approaches are independent in terms of knowledge acquisition for question summarization. In the future, we aim to investigate methods and architectures allowing to combine both approaches to further improve the summarization of consumer health questions. 

\section*{Acknowledgments}
This work was supported by the intramural research program at the U.S. National Library of Medicine, National Institutes of Health.

 


 \bibliographystyle{plainnat} 
 \bibliography{cas-refs}





\end{document}